\documentclass[
twocolumn
]{ceurart}

\usepackage{listings}
\usepackage{paralist}
\usepackage{caption}

\newcommand\wordblank[1]{\underline{\phantom{#1}}}

\begin{document}

\copyrightyear{2022}
\copyrightclause{Copyright for this paper by its authors.
  Use permitted under Creative Commons License Attribution 4.0
  International (CC BY 4.0).}

\conference{RecSys in HR'22: The 2nd Workshop on Recommender Systems for Human Resources, in conjunction with the 16th ACM Conference on Recommender Systems, September 18--23, 2022, Seattle, USA.}

\title{Flexible Job Classification with Zero-Shot Learning}

\author{Thom Lake}[%
email=tlake@indeed.com
]
\address{Indeed}

\begin{abstract}
    Using a taxonomy to organize information requires classifying objects (documents, images, etc) with appropriate taxonomic classes.
    The flexible nature of zero-shot learning is appealing for this task because it allows classifiers to naturally adapt to taxonomy modifications.
    This work studies zero-shot multi-label document classification with fine-tuned language models under realistic taxonomy expansion scenarios in the human resource domain.
    Experiments show that zero-shot learning can be highly effective in this setting.
    When controlling for training data budget, zero-shot classifiers achieve a 12\% relative increase in macro-AP when compared to a traditional multi-label classifier trained on all classes.
    Counterintuitively, these results suggest in some settings it would be preferable to adopt zero-shot techniques and spend resources annotating more documents with an incomplete set of classes, rather than spreading the labeling budget uniformly over all classes and using traditional classification techniques.
    Additional experiments demonstrate that adopting the well-known filter/re-rank decomposition from the recommender systems literature can significantly reduce the computational burden of high-performance zero-shot classifiers, empirically resulting in a 98\% reduction in computational overhead for only a 2\% relative decrease in performance.
    The evidence presented here demonstrates that zero-shot learning has the potential to significantly increase the flexibility of taxonomies and highlights directions for future research.
\end{abstract}

\begin{keywords}
  Taxonomy \sep
  zero-shot learning \sep
  multi-label classification \sep
  natural language processing
\end{keywords}

\maketitle

\section{Introduction}
Taxonomies used to organize information must frequently be adapted to reflect external changes such as the introduction of new markets, the creation of specialized segments, or the addition of new features.
This is especially true in the human resource (HR) domain, where new job, skill, and license categories must be created to accommodate a constantly evolving marketplace.
Unfortunately, the techniques commonly used to label real-world objects (documents, images, etc) with taxonomy classes are tightly coupled to the specific set of classes available when the classification system is developed.
In order to add \textit{new classes}, rule-based systems \cite{chiticariu2013rule, kejriwal2019expertRules} require the creation of new rules, and supervised machine learning techniques \cite{cucerzan2007entities, karadeniz2019linkingBio, lee2013attribute, ghani2006productAttribute} require labeling data with the new classes and training a new model.
These requirements make operationalizing modifications of the underlying taxonomy cumbersome.

Unlike traditional supervised classification techniques, zero-shot learning (ZSL) techniques are able to generalize to new classes with minimal guidance  \cite{larochelle2008zeroData, chang2008Dataless}.
Applying ZSL to taxonomic classification has the potential to increase the flexibility of organizational data structures while retaining the performance benefits of machine learning techniques.

Within this context, this work empirically studies the performance of ZSL techniques for document classification in the HR domain.
Experiments designed to simulate realistic taxonomy expansion scenarios show that ZSL is highly effective, outperforming standard supervised classifiers in low-resource settings.
Further experiments demonstrate that adopting well-known techniques can significantly reduce the computational overhead of high-performance zero-shot classifiers.

\section{Related Work}

There is a large body of previous work on ZSL \cite{larochelle2008zeroData, chang2008Dataless, xian2017zeroshotGoodBad, chen2021zsKnowledgeSurvey}.
Early work in the computer vision domain \cite{socher2013zeroCrossModal} represented classes with pre-trained word embeddings \cite{mikolov2013word2vec} and trained models to align them with image embeddings in a shared vector space.
Much of the subsequent work in ZSL has followed a similar embedding-based approach \cite{romera2015embarrassingly, xian2016latentZeroshot, qiao2016zsCVFromDocs, zhang2017learningEmbedding}.

A common assumption in ZSL is that the set train and test classes are disjoint.
Noting that this is somewhat unrealistic, \cite{chao2016generalizedIntro} proposed generalized zero-shot learning (GZSL), which assumes training classes are a strict subset of test classes \cite{liu2018generalizedZeroshot, pourpanah2020zsGeneralizedReview}.
As this work is primarily concerned with classifiers that can adapt to a changing taxonomy, experiments are conducted within the GZSL framework.

While there has been less explicit research on ZSL for NLP, as noted by \cite{li2018zsTextCNN}, most techniques for ad-hoc document retrieval \cite{robertson2009bm25, wei2006ldaAdHoc} can be leveraged for zero-shot document classification by treating the labels as queries.
In \cite{pushp2017trainOnce}, a standard classifier was applied to a combined representation of a document and label, produced with word embeddings or LSTMs \cite{hochreiter1997lstm}.
\cite{li2018zsTextCNN} apply convolution neural networks \cite{kim2014textCNN} over features derived from interactions between token and class embeddings.

Following the rise of transfer learning via fine-tuning for NLP \cite{howard2018ulmfit, devlin2019bert}, recent approaches to zero-shot document class classification have adopted similar techniques.
In \cite{yin2019zeroshotEntailment} zero-shot document classification was formulated as an entailment task.
Pre-trained language models were either fine-tuned on a dataset containing a subset of classes, or datasets for natural language inference (NLI) \cite{adina2018mnli}.
An identical entailment formulation was used in \cite{halder2020taskAware}, which studied zero-shot transfer between datasets.
Pre-trained language models were also used for zero-shot document classification in \cite{schick2021exploitingCloze}, which explored the use of cloze-style templates for zero-shot and few-shot document classification.

Autoregressive neural language models have been shown to possess some ZSL capabilities with proper prompting \cite{radford2019gpt2}.
Significantly larger models have improved these results \cite{brown2020GPT3}.
However, the computational demands of such large models make them unsuitable for most practical applications.

The benefit of fine-tuning for entailment-based ZSL was studied in \cite{ma2021issuesEntailment}.
Their experiments showed fine-tuning on generic NLI datasets often results in worse ZSL performance and hypothesize this is due to models exploiting lexical patterns and other spurious statistical cues \cite{feng2019misleadingFailure, niven2019probingFailure}.
Experimental results presented here complement those in \cite{ma2021issuesEntailment}, suggesting their observations do not apply when even a small amount of task-specific training data is available.

The closely related work of \cite{chalkidis2020empiricalMultilabel} also studied GZSL for multi-label text classification. 
Their focus was on understanding the role of incorporating knowledge of the hierarchical label structure into models in both the few-shot and zero-shot settings.
Instead, the work presented here specifically designs experiments to better understand the ability of standard GZSL techniques to generalize in realistic zero-shot settings when orders of magnitude less background training data are available.

\begin{figure*}[h]
    \centering
    \includegraphics[width=1\linewidth]{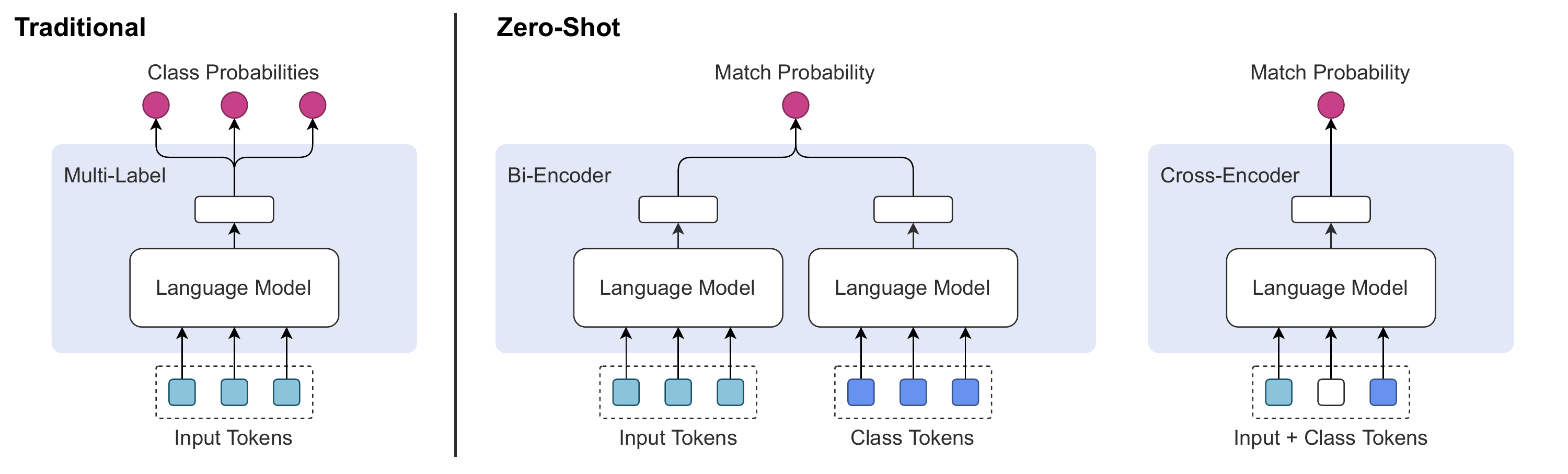}
    \caption{Graphical representation of models used in experiments.
    Traditional multi-label classifiers (left) output a probability for each class.
    Zero-shot classifiers (right) model compatibility between an input and class description.}
    \label{fig:models}
\end{figure*}

\section{Problem Formulation}

Taxonomy classification is formulated in terms of a multi-label text classification problem.
Let $Y$ be a set of classes, $x_i \in X$ a document, and $\mathbf{y}_i \in \{0,1\}^{|Y|}$ a corresponding binary label vector where $y_{ij} = 1$ if document $x_i$ is labeled with class $j$ and $0$ otherwise.
A common probabilistic approach to multi-label text classification \cite{murphy2012} is to assume conditional independence among labels,
\begin{align}
    \label{eq:formulation}
    p(\mathbf{y}_i \mid x_i) = \prod_j p(y_{ij} \mid x_i) = \prod_j q_{ij}^{y_{ij}}(1 - q_{ij})^{1 - y_{ij}},
\end{align}
and approximate the parameters of the conditional Bernoulli distributions, $0 \le q_{ij} \le 1$, using some model.
A common choice is $q_{ij} \approx \sigma(r_{ij}) = (1 + e^{-r_{ij}})^{-1}$, where
\begin{align}
    \label{eq:multi-label}
    r_{ij} = \mathbf{w}_j^T f_\theta(x_i),
\end{align}
$\mathbf{w}_j \in \mathbb{R}^d$ is a vector of parameters, and $f_\theta \colon X \rightarrow \mathbb{R}^d$ is a function with parameters $\theta$, e.g., a transformer neural network \cite{vaswani2017attention}.
In the remainder, the above is simply referred to as the standard \textit{multi-label} model.

Because each class $j$ is associated with a distinct vector of parameters $\mathbf{w}_j$ in (\ref{eq:multi-label}), the multi-label model is unable to generalize to classes not observed during training.
To side-step this issue, ZSL assumes the existence of textual class descriptions $z_j \in X$ for each class $j \in Y$ which can be leveraged to break the explicit dependency between model parameters and classes.
This work considers two standard architectures from the literature \cite{humeau2019poly}, described below and depicted graphically in Figure \ref{fig:models}, which can incorporate class descriptions. 
Models are designed to be relatively simple, reflective of common best practices, and as similar as possible to avoid confounding and draw clear inferences about general performance patterns.

\textbf{Bi-Encoder:} This model replaces the vector $\mathbf{w}_j$ with the output of an additional parameterized function taking class descriptions as input,
\begin{align*}
    r_{ij} = f_{\theta_1}(z_j)^T f_{\theta_2}(x_i).
\end{align*}

\textbf{Cross-Encoder:} A parameterized function that takes as input a concatenated document and class description (denoted by $\sqcup$). The model has a single additional parameter vector $\mathbf{w} \in \mathbb{R}^d$,
\begin{align*}
    r_{ij} &= \mathbf{w}^T f_\theta(x_i \sqcup z_j).
\end{align*}

\subsection{Loss}

Given a dataset $D = \{(x_1, \mathbf{y}_1), \ldots, (x_{|D|}, \mathbf{y}_{|D|})\}$, model parameters can be optimized by minimizing negative log-likelihood
\[
    \mathcal{L}(D) = |D|^{-1} \sum_i \ell(i),
\]
where
\begin{align}
    \ell(i) &= -\sum_j (y_{ij} \log\sigma(r_{ij}) + \left(1 - y_{ij}\right) \log\left(1 - \sigma(r_{ij})\right) \label{eq:loss}
\end{align}
Due to zero-shot approaches conditioning on class descriptions, computing the sum over each class in Equation (\ref{eq:loss}) requires $|Y|$ forward passes through the model. 
This results in significant computational overhead when training. 
To alleviate this issue, the commonly used negative sampling \cite{mikolov2013word2vec} strategy is used to approximate the loss $\ell(\cdot)$, 
\begin{align}
    \hat{\ell}(i) &= -\log\frac{e^{r_{ij}}}{e^{r_{ij}} + \sum_{i'} e^{r_{i'j}} + \sum_{j'} e^{r_{ij'}}} \label{eq:loss-ns}
\end{align}
where $i', j, j'$ are uniformly sample such that $y_{ij} = 1$ and $y_{i'j} = y_{ij'} = 0$.
The number of negative documents $i'$ and classes $j'$ are treated as hyper-parameters.
Initial experiments also explored a Bernoulli rather than a categorical version of $\hat{\ell}(\cdot)$ but found the categorical version performed better.

\section{Experiments}

Experiments are designed to simulate real-world taxonomy expansion driven by domain experts.
At a high level, all experiments follow the same process.
\begin{enumerate}
    \item Modify or remove classes to obtain the \textbf{Source Taxonomy}.
    Critically, this is done in a way that incorporates the underlying structure of the taxonomy to ensure coherent modifications, rather than simply removing classes at random.
    \item Train classifiers using a dataset of instances labeled with classes from the \textbf{Source Taxonomy}.
    \item Expand the \textbf{Source Taxonomy} by undoing the modifications from Step 1 to obtain the \textbf{Target Taxonomy}.
    \item Evaluate classifiers on a new dataset of instances labeled with classes from the \textbf{Target Taxonomy}.
\end{enumerate}
Details of the taxonomy, datasets, and expansion types used in this work are given below.

\subsection{Indeed Occupations}

Indeed's internal U.S occupation taxonomy was used as a representative source of structured knowledge.
The taxonomy contains over a thousand occupations arranged hierarchically in a forest-like directed acyclic graph (DAG), with root nodes being general occupations, \textit{Healthcare Occupations}, and leaf nodes being the most specific, \textit{Nurse Practitioners}.
In addition to their placement within the hierarchy, occupations are also associated with a natural language \textbf{name} and \textbf{definition}.
Data formats are given in Table \ref{tab:documents}.
\begin{table}[h]
    \caption{The data representations used in this work.
    Jobs and occupations are converted to strings composed of multiple fields.}
    \label{tab:documents}
    \begin{tabular}{ll}
        \toprule
        Object & Text \\
        \midrule
        Job & Title: \wordblank{foo}, Employer: \wordblank{foo}, Description: \wordblank{foo} \\
        Occupation & Name: \wordblank{foo}, Definition: \wordblank{foo} \\
        \bottomrule
    \end{tabular}
\end{table}

Each job posted on Indeed is labeled with one or more occupations.
The number of jobs per occupation for evaluation data is given in Table \ref{tab:test-data}.
Jobs were selected using stratified sampling by occupation.
In particular, for each occupation $N$ jobs labeled with that occupation were randomly sampled without replacement.
It should be noted that since jobs can be labeled with multiple occupations, this sampling strategy only guarantees datasets contain at least $N$ jobs per occupation, not that there are exactly $N$ jobs per occupation.
The same procedure was used to sample disjoints subsets of jobs for training, validation, and testing.

\begin{table}[h]
    \caption{Test jobs by numbers of labels.
    Five jobs were sampled for each occupation for evaluation.
    }
    \label{tab:test-data}
    \begin{tabular}{r | rr}
        \toprule
        Labels & Jobs & Percent \\
        \midrule
        1 & 2,527 & 55\% \\
        2 & 1,567 & 34\% \\
        3 & 393 & 9\% \\
        4 & 68 & 1\% \\
        \midrule
        \textbf{Total} & 4,555  & 100\% \\
        \bottomrule
    \end{tabular}
\end{table}

\begin{figure}[h]
    \centering
    \includegraphics[width=1\linewidth]{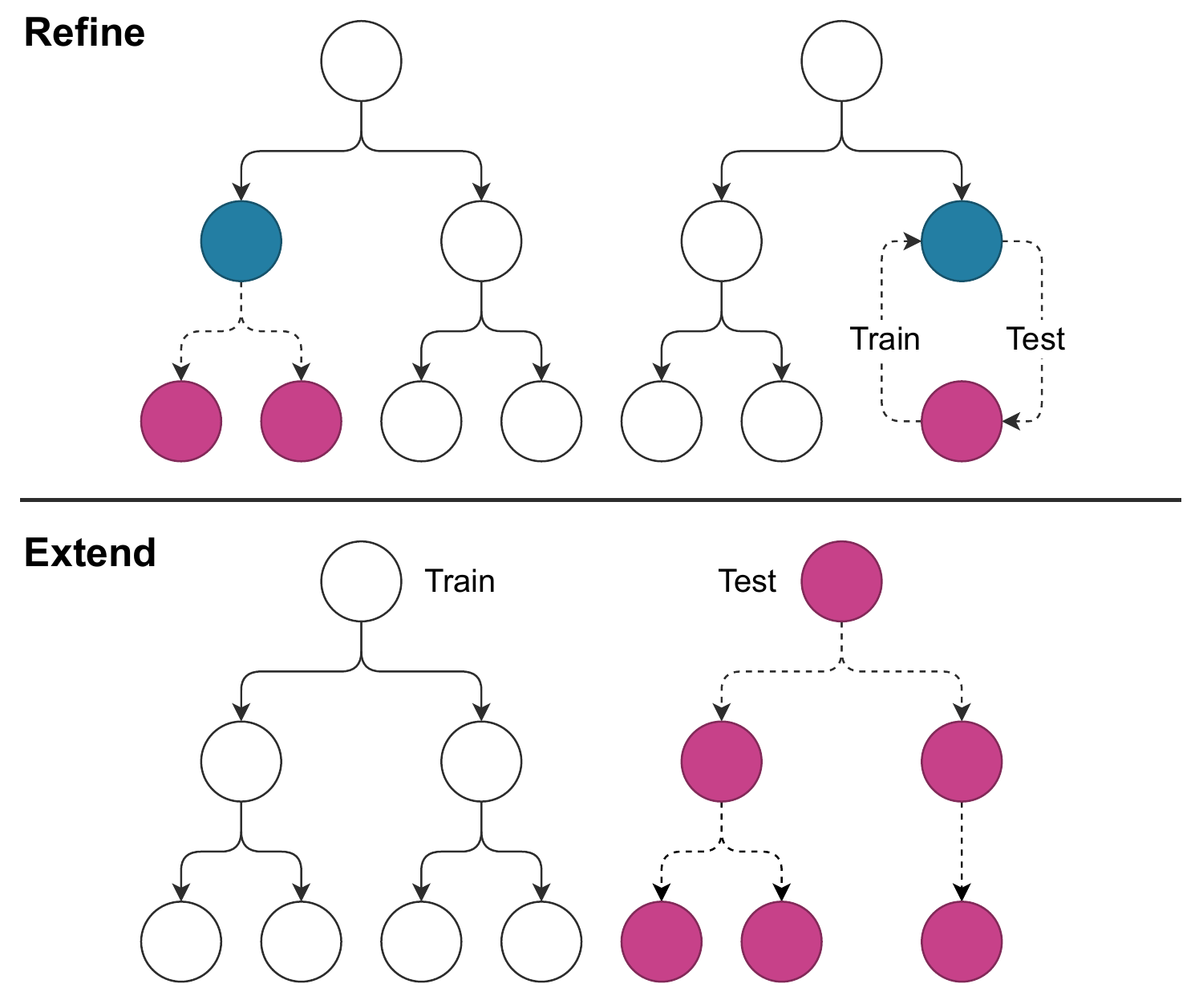}
    \caption{Graphical representation of \textbf{Refine} (top) and \textbf{Extend} (bottom) taxonomy expansion operations.
    Each node represents a class.
    Models are evaluated on all classes.
    White and teal classes are observed during training.
    Magenta classes are not observed during training.
    Teal classes replace their children during training.}
    \label{fig:experiments}
\end{figure}

\subsection{Expansion Operations}

The two taxonomy expansion operations considered are described below and depicted graphically in Figure \ref{fig:experiments}.

\textbf{Refine:} This setting simulates the scenario where a subset of leaf classes are subdivided into more fine-grained classes.
This sort of refinement can occur when gaps in the taxonomy surface after use, or in situations when the set of initial classes naturally diversifies over time.
For example, an academic field of study may subdivide into more specialized subfields as it matures.
Zero-shot classifiers in this setting must generalize to classes that are more specific versions of those encountered during training. 

To construct datasets in this setting, a random leaf class is selected.
Any appearances of this class or its siblings are replaced with the parent class.
This process is repeated until a fixed percentage of leaf classes have been replaced. 

\textbf{Extend:} This setting simulates the scenario where a set of classes are added from an unrelated domain.
This situation can occur when new use cases surface that require classes that were not previously necessary.
For example, if an e-commerce company that had historically only sold goods like household items and clothing began offering groceries, the previous product taxonomy would not be useful for organizing the new items.
Zero-shot classifiers in this setting must generalize to classes that are significantly different from those encountered during training.

To construct datasets in this setting, a random root class is selected.
Any appearances of this class or its descendants are removed.
This process is repeated until a fixed percentage of classes have been removed.
At the end of the process, any document that no longer has any labels is removed from the training dataset.

\subsection{Evaluation}

Performance is evaluated in terms of a model's ability to rank relevant classes for a particular documents, and rank documents with respect to a class. 
In both cases, average precision (AP) is used to measure the quality of a predicted ordering relative to ground truth labels. 
The difference is whether AP is computed for all labels and averaged over documents, typically referred to as label ranking average Precision (LRAP) \cite{tsoumakas2009lrap}, or computed for all documents and averaged over labels, typically referred to as macro-AP.
Formally, for matrices $\mathbf{Y} \in \{0,1\}^{|D| \times |Y|}$ of ground truth binary labels and $\mathbf{R} \in \mathbb{R}^{|D| \times |Y|}$ of predicted scores, then
\begin{align*}
    \text{LRAP} &= |D|^{-1} \sum_i \text{AP}\left(\mathbf{Y}_{i,:}, \mathbf{R}_{i,:}\right) \\
    \text{macro-AP} &= |Y|^{-1} \sum_j \text{AP}\left(\mathbf{Y}_{:,j}, \mathbf{R}_{:,j}\right)
\end{align*}
where for vectors $\mathbf{y} \in \{0,1\}^d$ and $\mathbf{r} \in \mathbb{R}^d$
\[
    \text{AP}(\mathbf{y}, \mathbf{r}) = \frac{1}{\sum_i y_i}\sum_i y_i \frac{|\{k \mid y_k = 1 \land r_k \ge r_i\}|}{|\{k \mid r_k \ge r_i\}|}.
\]

\subsection{Training Details}

Following modern practices in NLP, models consist of a pre-trained transformer \cite{vaswani2017attention} backbone which is fine-tuned \cite{howard2018ulmfit, devlin2019bert} along with any additional parameters.
All models use BERT-base \cite{devlin2019bert} as a backbone language model.
Hyper-parameters were manually tuned on a small subset of the training data using the multi-label model and fixed for all models and experiments.
The Adam \cite{kingma2015adam} optimizer was used with a learning rate of 2e-5 for pre-trained parameters and 2e-4 for randomly initialized parameters. 
Learning rate warm-up was applied for the first 10\% of the updates and then linearly decayed to zero.
The maximum gradient norm was clipped to 10 \cite{zhang2019gradientClipping}.
All models are trained for 20 epochs with a batch size of 64.
Models are evaluated after each epoch and the final model is selected based on the LRAP on the validation dataset.
The bi-encoder and cross-encoder models were trained using negative sampling with 8 negative classes and 4 negative inputs per positive training document (Equation \ref{eq:loss-ns}).
Experiments utilized the PyTorch \cite{paszke2019pytorch} and Huggingface Transformers \cite{wolf2020transformers} libraries.
All hyper-parameters not listed explicitly above are left to their default values.
Experiments were conducted using a single NVIDIA Tesla V100 GPU with 16GB of memory.

\section{Results}

\subsection{Generalizing to Novel Classes}

\begin{table*}[h]
\caption{LRAP and Macro-AP for by model, class coverage, minimum documents per class, and number of training documents in the extend setting.
Models denoted by $\dagger$ do not observe any task-specific training data.
}
\label{tab:multi-setting}
\begin{tabular}{lrrrrr}
\toprule
Model    & Class Coverage   &   Documents Per Class &   Documents &   LRAP &   macro-AP \\
\midrule
 Multi-Label   & 100\%               &                    3 &        2733 &  0.569 &      0.496 \\
 Multi-Label   & 50\%                &                    5 &        2500 &  0.294 &      0.249 \\
 Bi-Encoder    & 50\%                &                    5 &        2500 &  0.362 &      0.349 \\
 Cross-Encoder & 50\%                &                    5 &        2500 &  \textbf{0.645} &      \textbf{0.553} \\
 \midrule
 Multi-Label   & 100\%               &                    4 &        3614 &  0.638 &      0.564 \\
 Multi-Label   & 75\%                &                    5 &        3628 &  0.493 &      0.438 \\
 Bi-Encoder    & 75\%                &                    5 &        3628 &  0.480 &      0.447 \\
 Cross-Encoder & 75\%                &                    5 &        3628 &  \textbf{0.654} &      \textbf{0.590} \\
 \midrule
 Multi-Label   & 100\%               &                    5 &        4555 &  \textbf{0.697} &      \textbf{0.635} \\
 Bi-Encoder    & 100\%               &                    5 &        4555 &  0.570 &      0.521 \\
 Cross-Encoder & 100\%               &                    5 &        4555 &  0.682 &      0.613 \\
 \midrule
 Cross-Encoder (NSP)\textsuperscript{$\dagger$} & - & - & - & 0.419 & 0.242 \\
 TF-IDF\textsuperscript{$\dagger$}        & - & - & - & 0.397 & 0.294  \\
\bottomrule
\end{tabular}
\end{table*}

\begin{figure}[h]
    \centering
    \includegraphics[width=1\linewidth]{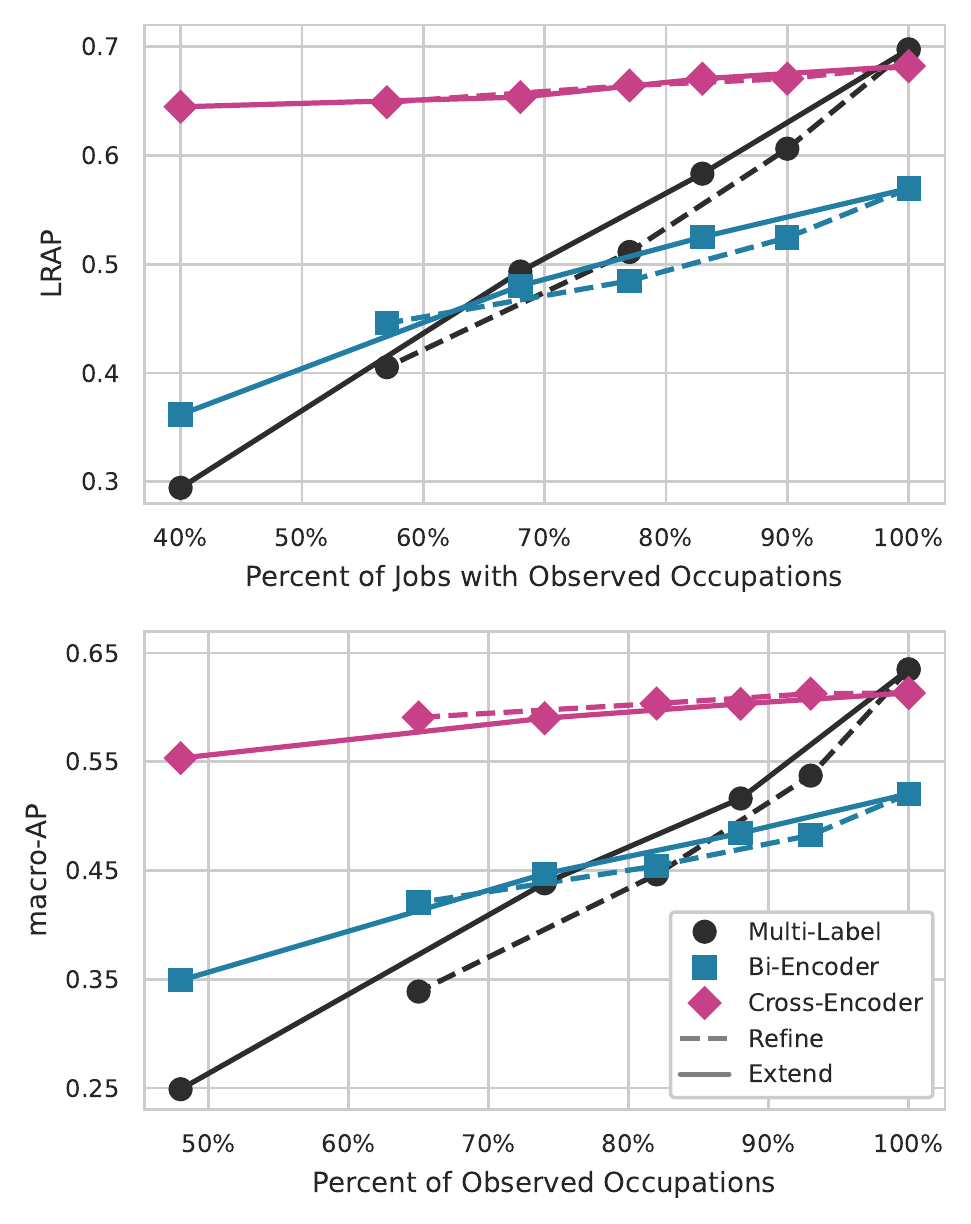}
    \caption{LRAP (top) and macro-AP (bottom) under different taxonomy expansion operations.
    Models are identified by color and symbol.
    Line styles reflect the expansion operation, with dashed lines for refinement and solid lines for extension.}
    \label{fig:performance-combined}
\end{figure}

Performance was evaluated for different percentages of observed classes during training (coverage) for both the refine and extend expansion operation.
LRAP and macro-AP are shown in Figure \ref{fig:performance-combined}.
The cross-encoder classifier was robust to both taxonomy refinement and expansion.
Minimal performance degradation was observed with decreasing coverage, even in settings where over 50\% of the classes are new and approximately 60\% of the jobs are labeled with a new occupation.
The bi-encoder performed significantly worse than the cross-encoder.
This observation is consistent with prior-work in the retrieval domain \cite{humeau2019poly, khattab2020colbert}.
However, the bi-encoder also suffered more performance degradation with decreasing coverage.
For example, the bi-encoder's macro-AP dropped by 36\% when 50\% of the classes are new (extend), whereas the macro-AP cross-encoder's only decreased by 5\%.
Performance of the multi-label classifier degraded rapidly as coverage deceased, as it is unable to generalize to classes not observed during training.

\subsection{Learning on a Budget}

Because the extend operation omits labels rather than relabeling them, zero-shot models had access to less training data in the previous experiments.
To better understand the trade-off between fine-tuning and ZSL, experiments were conducted which controlled for the amount of data available for training.
In particular, multi-label classifiers were trained on datasets where the number of documents was similar to ZSL approaches, but fewer documents per class are observed.
Full results are presented in Table \ref{tab:multi-setting}.
The ZSL cross-encoder with 50\% coverage and five documents per class resulted in a 13\% relative increase in LRAP over the multi-label classifier with 100\% coverage and three documents per class (similar training set size).
This result was unexpected, as it suggests that given a small document labeling budget (\textless 4K here), in some settings it would be preferable to adopt ZSL and spend resources annotating more documents with an incomplete set of classes, rather than spreading the labeling budget uniformly over all classes and using traditional classifiers.

Further analysis of zero-shot performance is given in Table \ref{tab:new-trees}, which presents macro-AP by root class for unobserved classes in the extend setting with 50\% coverage.
Despite not being previously exposed to any classes from these domains, in all cased the cross-encoder outperformed the multi-label classifier explicitly trained on these classes.

\begin{table*}[h]
\caption{Zero-shot Macro-AP for novel domains in the challenging extend scenario with 50\% class coverage.
$\dagger$ Because the multi-label classifier is not capable of zero-shot generalization, it is trained with 100\% class coverage, but fewer documents per class.}
\label{tab:new-trees}
\begin{tabular}{lrrrr}
\toprule
 Domain                                      &   Classes & Bi-Encoder &   Cross-Encoder &   Multi-Label\textsuperscript{$\dagger$}\\
\midrule
 Personal Service                          &         28 &        0.273 &           0.642 &             0.590 \\
 Food \& Beverage                           &         25 &        0.245 &           0.619 &             0.555 \\
 Cleaning \& Grounds Maintenance            &         25 &        0.277 &           0.584 &             0.563 \\
 Marketing, Advertising \& Public Relations &         28 &        0.241 &           0.579 &             0.541 \\
 Repair, Maintenance \& Installation        &         34 &        0.276 &           0.533 &             0.494 \\
 Healthcare                                &        156 &        0.250 &           0.532 &             0.584 \\
 Protective \& Security                     &         27 &        0.302 &           0.529 &             0.509 \\
 Construction \& Extraction                 &         54 &        0.265 &           0.527 &             0.465 \\
 Architecture \& Engineering                &         36 &        0.207 &           0.474 &             0.399 \\
 Sales, Retail \& Customer Support          &         31 &        0.244 &           0.472 &             0.453 \\
 Supply Chain \& Logistics                  &         32 &        0.243 &           0.457 &             0.435 \\
 \midrule
 New Classes                                &        478 &        0.251 &          \textbf{0.534} &             0.523 \\
 Old Classes                                      &        433 &        0.457 &    \textbf{0.575} &             0.465 \\
 All Classes                                       &        911 &        0.349 &   \textbf{0.553} &             0.496 \\
 \midrule
 Training Documents && 2500 & 2500 & 2733 \\
 Documents Per Class && 5 & 5 & 3 \\
 Class Coverage && 50\% & 50\% & 100\% \\
\bottomrule
\end{tabular}
\end{table*}

\subsection{Efficient Zero-Shot Inference}
\label{subsec:efficient}

\begin{figure}[h]
    \centering
    \includegraphics[width=1\linewidth]{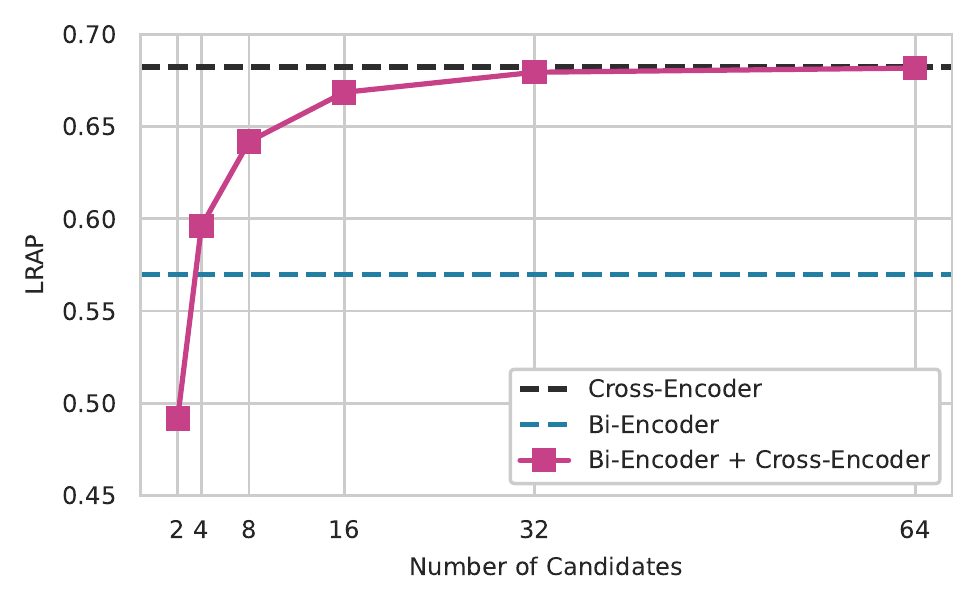}
    \caption{LRAP for two-phase zero-shot classification for candidates set sizes from 2 to 64.
    Dashed lines depict the performance of standalone models.}
    \label{fig:reranking}
\end{figure}

As noted previously, there is a significant computational cost associated with training the transformer-based zero-shot learners due to the need to process each label for each document.
While this cost can be amortized for the bi-encoder at inference time by pre-computing label embeddings, this is not possible for the cross-encoder architecture.
Several works explore the architecture space between bi-encoders and cross-encoders to obtain a better trade-off between performance and latency \cite{humeau2019poly, khattab2020colbert}.
A simpler technique was explored in this work inspired by the common decomposition of recommender systems into separate candidate retrieval and re-ranking \cite{covington2016youtube} phases.

In the first phase, the more efficient bi-encoder was used to identify a small subset of potentially relevant candidate classes.
This smaller set of candidates was then evaluated with the more computationally demanding, but higher performance cross-encoder.
Classes not selected in the first phase were implicitly assumed to receive a score of zero.
Results are shown in Figure \ref{fig:reranking} for candidate set sizes from 2 to 64.
Scoring only 16 candidates resulted in a small drop in LRAP (-2\%) while resulting in a nearly 98\% reduction in computational overhead.

\section{Conclusion and Future Work}
\label{sec:conclusion}
Taxonomies are widely used to organize knowledge and can easily incorporate important information from domain experts that may be difficult to obtain in a purely automated fashion.
However, the ability to associate classes with real-world classes can be a bottleneck for the rapid expansion of taxonomies.
Experiments presented here demonstrate that modern zero-shot classification techniques can sidestep this issue by classifying objects with novel classes using only minimal human guidance.

Better understanding and overcoming the failure modes of the bi-encoder architecture would result in more efficient systems capable of scaling larger taxonomies, either as stand-alone systems or as part of a multi-phase such as that described in Section \ref{subsec:efficient}.
Related work in the retrieval setting suggests adopting pretext \cite{jing2020selfPretext} tasks that are better aligned with the downstream task of interest could alleviate these issues \cite{chang2019embeddingBasedRetrieval}.
Alternatively, more elaborate negative sampling strategies \cite{weston2011wsabie, zhan2021hardNegatives} could improve both zero-shot techniques studied in this work, and close any observed gaps between zero-shot learners and traditional classifiers.
Future work should explore zero-shot capabilities in more sophisticated knowledge bases (ontologies, knowledge graphs, etc), a larger variety of class types, and different domains.
Lastly, further experimentation is needed to fully explain observed differences between the results presented here and those in \cite{ma2021issuesEntailment} in order to better understand the success and failure modes of entailment-based ZSL.

\begin{acknowledgments}
  Valuable insights, suggestions, and feedback was provided by numerous individuals at Indeed.
  The author would especially like to thank Suyi Tu, Josh Levy, Ethan Handel, Arvi Sreenivasan, and Donal McMahon.
\end{acknowledgments}

\bibliography{zero-shot-job-classification}


\end{document}